\pdfoutput=1
\PassOptionsToPackage{numbers}{natbib}
\documentclass[11pt]{article}

\usepackage[preprint]{acl}

\usepackage{times}
\usepackage{latexsym}

\usepackage[T1]{fontenc}

\usepackage[utf8]{inputenc}

\usepackage{microtype}

\usepackage{inconsolata}

\usepackage{graphicx}

\usepackage{amsmath}
\usepackage{algorithm}
\usepackage{algpseudocode}
\usepackage{multirow} 
\usepackage{xcolor}
\usepackage{colortbl}
\usepackage{booktabs}

\usepackage{setspace}
\usepackage{subfigure}
\usepackage{cleveref}
\usepackage[numbers]{natbib}
\usepackage{amsfonts}

\title{Enhancing point cloud analysis via  neighbor aggregation correction based on cross-stage structure correlation}

\author{
 \textbf{Jiaqi Shi\textsuperscript{1}},
 \textbf{Jin Xiao\textsuperscript{1}},
 \textbf{Xiaoguang Hu\textsuperscript{1}},
 \textbf{Boyang Song\textsuperscript{1}}
 \textbf{Hao Jiang\textsuperscript{1}}
 \textbf{Tianyou Chen\textsuperscript{2}}
 \textbf{Baochang Zhang\textsuperscript{3}}
 \\
  \textsuperscript{1}School of Automation Science and Electrical Engineering, Beihang University \\
  \textsuperscript{2}Chinese Aeronautical Radio Electronics Research Institute  \\
  \textsuperscript{3}School of Artificial Intelligence, Beihang University \\
  \\
}

\begin{document}
\maketitle
\begin{abstract}
Point cloud analysis is the cornerstone of many downstream tasks, among which aggregating local structures is the basis for understanding point cloud data. While numerous works aggregate neighbor using three-dimensional relative coordinates, there are irrelevant point interference and feature hierarchy gap problems due to the limitation of local coordinates.
Although some works address this limitation by refining spatial description though explicit modeling of cross-stage structure, these enhancement methods based on direct geometric structure encoding have problems of high computational overhead and noise sensitivity. 
To overcome these problems, we propose the Point Distribution Set Abstraction module (PDSA) that utilizes the correlation in the high-dimensional space to correct the feature distribution during aggregation, which improves the computational efficiency and robustness. PDSA distinguishes the point correlation based on a lightweight cross-stage structural descriptor, and enhances structural homogeneity by reducing the variance of the neighbor feature matrix and increasing classes separability though long-distance modeling. Additionally, we introducing a key point mechanism to optimize the computational overhead.
The experimental result on semantic segmentation and classification tasks based on different baselines verify the generalization of the method we proposed, and achieve significant performance improvement with less parameter cost. The corresponding ablation and visualization results demonstrate the effectiveness and rationality of our method. The code and training weight is available at: \url{https://github.com/AGENT9717/PointDistribution}
\end{abstract}    
\section{Introduction}
\label{sec:intro}

Recently, three-dimensional sensors represented by LiDAR have been applied and developed at a high speed in many fields such as remote sensing, autonomous driving and smart city \cite{pointapply01,pointapply02,pointapply03, zhu2024tmsdnet,xiao2024human}. 
These applications \cite{pointcloudsurvey} focus on how to understand the real world from point cloud data. Therefore, 3D point cloud analysis becomes the cornerstone of many downstream tasks.

After the pioneer work PointNet \cite{pointnet}, PointNet++ \cite{pointnet++} proposes the Point Set Abstraction (SA) module to extract point neighbor structure, which enables PointNet \cite{pointnet} to capture local details and establishes PointNet++ \cite{pointnet++} as a foundational framework for subsequent research \cite{pointnext,pointmeta,pointvector,pointnat}; the fundamental extracting structure module during down-sampling in many works remains similar to the original SA module: sampling and grouping point cloud neighbor, encoding local three-dimensional relative coordinates in neighbor, and aggregating features through pooling operations.

\begin{figure}[ht]
    \centering
    \vspace{0cm} 
    \subfigure[Refine the description of low-dimensional geometric structures]{
        \begin{minipage}[b]{1.0\linewidth}
            \centering
            \label{fig:refining}
            \includegraphics[width=1.0\linewidth]{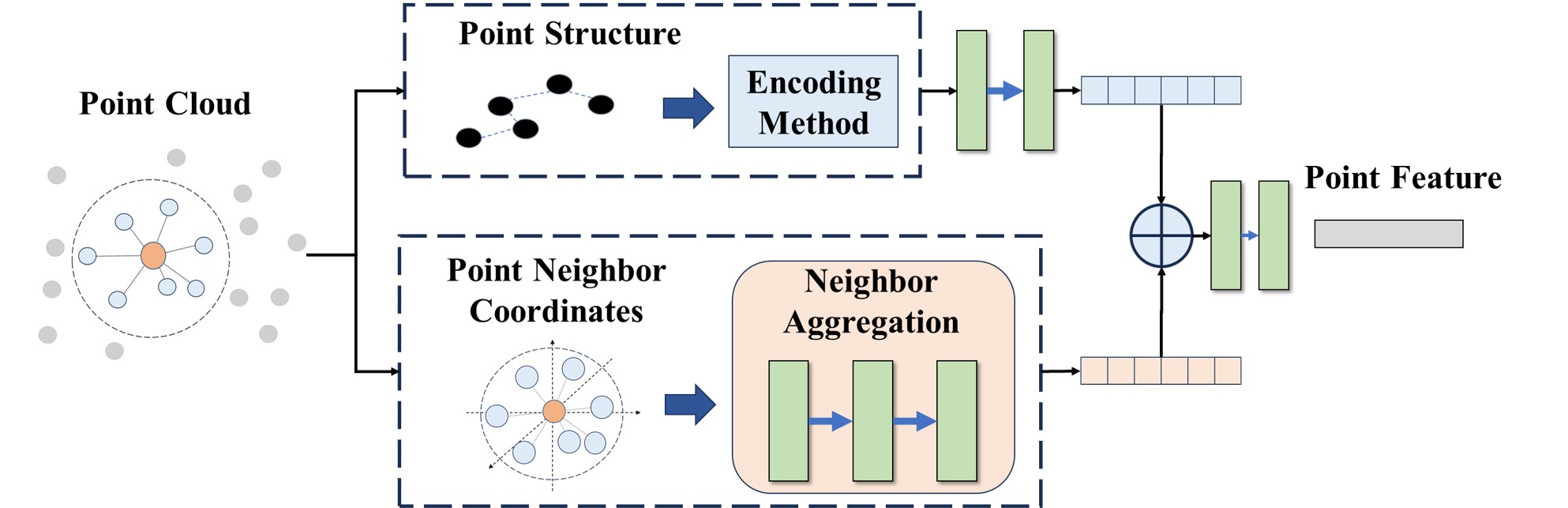}
        \end{minipage}
    }
    \subfigure[Correct the feature distribution in high-dimensional space]{
        \begin{minipage}[b]{1.0\linewidth}
            \centering
            \label{fig:correcting}
            \includegraphics[width=1.0\linewidth]{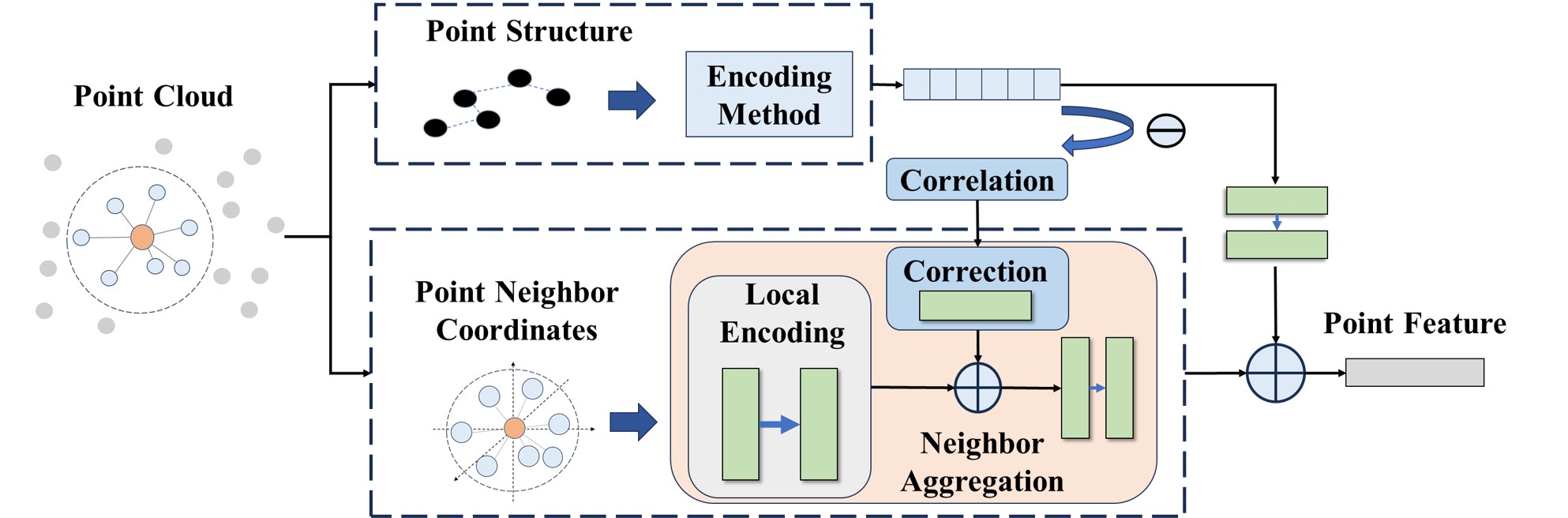}
        \end{minipage}
    }
    \caption{Illustrations of improving neighbor aggregation by cross-stage structure encoding.}
    \label{fig2}
    \vspace{-8pt} 
\end{figure}

However, there are some problems in such feature aggregation process which is based on local relative coordinates:
First, due to the random sampling characteristic of the grouper (e.g. ball query or K-NN), the sampled point neighbor may contain the irrelevant point that will affect the generalization of the network. Second, three-dimensional relative coordinates is difficult to accurately describe relative relationships for point cloud data, which is typical manifold data in the non-Euclidean space \cite{x-3d}. 
Further, the hierarchical gap between local coordinate and high-level semantic features may lead to distribution confusion problems in high-dimensional spaces among the encoded features.

To address the above problems, some work have explored to explicitly embed cross-stage spatial feature to refine neighbor description, such as: PointHop \cite{pointhop} and X-3D \cite{x-3d}, which encode spatial octant centroids around each point for classification and irrelevant point denoising; PointNN \cite{pointNN}, employing trigonometric position encoding of neighboring points 3D coordinates for category prediction; DDGCN \cite{chen2023ddgcn} utilizing trigonometric direction-distribution encoding of adjacent points to enhance feature representations; SGG-Nets \cite{zhu2025sgg} proposes a Spherical Geometry Descriptor to enhance the rotation-invariance of the network by refining point feature describing.

While existing cross-stage structure encoding methods can improve neighbor feature extraction accuracy though refining structure describing as shown in Fig.\ref{fig:refining}, there are two fundamental limitations: (1) These methods improve the feature extraction by refining the description of the geometric space structure. Therefore, high-dimensional encoding is required to avoid information loss, which will bring huge computational overhead. (2) The modeling based on low-dimensional geometric structure needs to additionally establish a new feature extraction channel, so feature fusion can only be performed at the end. This segregated architecture induces feature-space discrepancies and amplifies noise interference.

Therefore, we explore a neighbor aggregation improving method based on the correlation of cross-stage structure, which improves the computational efficiency and robustness by correcting the distribution of neighbor features in the high-dimensional space rather than refining the geometric structure modeling. Our method is shown in Fig.\ref{fig:correcting}, which enhances structural coherence in aggregated representation at neighbor aggregating.

Specifically, we first proposed a lightweight cross-stage structure descriptor (LCSD) for describing point-wise correlation, which can describe the distribution of each point and the overall distribution of the neighbor across stages, and thus can simultaneously support the distinction of irrelevant points and the refinement of feature descriptions;
Further, we propose two correction operations during neighbor aggregation: (1) correction for denoising irrelevant points (CDIP), which minimize feature distribution variance for improving neighbor homogeneity based on the correlation of LCSD in the high-dimensional space; (2) correction for improving classes separability (CICS), which models long-range contextual dependencies based on LCSD for improving the separability between features; furthermore, we perform computational optimization for large scene point clouds through key point selection.

Combined with the above improvements, we proposed an improved neighbor aggregation module named as Point-
Distribution Set Abstraction module (PDSA). Based on PDSA module, we propose a novel neural network named PointDistribution for point cloud analysis, which can be built in variant versions with different local feature extraction modules. Experimental results demonstrate that our method effectively enhances the performance of PointNet series work while validating its effectiveness and generalization capability. The improved models achieve state-of-the-art performance levels.

Overall, our contributions can be summarized as follows:

1) We analyze the noise problems inherent in extracting neighbor structure based on local relative coordinate (exemplified by the Set Abstraction module) from two aspects: irrelevant point interference and feature hierarchy gap, and presented an improved method from the perspective of feature spatial distribution: PDSA.

2) We design a new lightweight cross-stage structure descriptor (LCSD), which strengthens the descriptive capability of spatial distributions and enables correlation analysis between individual points during neighbor aggregation.

3) Based on LCSD, we propose the correction for denoising irrelevant points (CDIP) operation which strengthens the generalization of network through enhancing the homogeneity of neighbor feature matrix.

4) Based on LCSD, we propose the correction for improving classes separability (CICS) operation which model long-range contextual for refining feature describing; further, we introduce a key point selection mechanism to optimize the computing cost of large-scale scenic point cloud.

5) Experimental evaluations with varying network configurations in different task datasets validate the effectiveness and generalization capability of our method, and the results achieve state-of-the-art performance. We also conduct a series of ablation and visualization experiments to verify the rationality of our method.



\section{Related Work}
\label{sec:related work}

\subsection{Point-based Network }
Due to the unordered structural characteristics, traditional image processing such as 2D convolution cannot be directly applied to point cloud analysis. To address this challenge, some approaches \cite{intro-cite-2d-01,intro-cite-2d-02,intro-cite-2d-03,intro-cite-2d-04,intro-cite-2d-05} project the 3D point cloud to the 2D plane for leverage existing 2-dimensional convolution architectures and some approaches \cite{intro-cite-voxel-01,intro-cite-voxel-02,intro-cite-voxel-03,intro-cite-voxel-04} discretize point clouds into 3D grids to enable efficient spatial feature extraction. However, these methods inevitably incur information loss during spatial transformation.

Point-based approaches \cite{pointnet,pointnet++,pointcnn,pointconv,pointnext,pointvector}
directly compute point cloud without extra voxel or projecting process. PointNet \cite{pointnet} first propose to use shared MLP to capture the global feature structure. PointNet++ \cite{pointnet++} then proposes a Point Set Abstraction module and construct a hierarchical structure to improve the representation of local information. Subsequently, numerous works focused on improving local information extraction. Graph-based methods \cite{grah-01-dynamic,graph-02,graph-03,li2021pointvgg,zhang2020local} uses the point feature and edge information between neighboring points to model local relationships. CNN-based methods \cite{pointcnn, pointconv, kpconv} propose different kinds of kernel points, and extract the local information based on the relative position between the point and the kernel point. Transformer-based \cite{pct,pointtransformerv1,pointtransformerv2,pointtransformerv3,li2025lcl-fda} methods use the correlation of points features and relative positions to extract the local information.

Recently, MLP-based method has shown the ability to rival the Transformer with simple architecture\cite{pointvector}, numerous works \cite{pointnext,pointmeta,pointvector,pointnat} have begun to refocus on PointNet++ which is the representative of the MLP architecture,  and consequently, most works in this series adopt neighbor feature aggregation structures similar to the SA module. We conduct an in-depth analysis of the noise interference problem existing in the neighbor aggregation process of SA module.

\subsection{Feature Aggregation on Point Cloud}
PointNet \cite{pointnet} and PointNet++ \cite{pointnet++} use max pooling as their global reduction function, which effectively captures geometric invariance through its computational simplicity. However, this approach risks discarding critical structural information during feature reduction.

Many works have begun to distinguish neighboring point correlations to improve the accuracy of neighbor aggregation.
RandLA-Net \cite{randla} utilizes channel attention pooling instead of max pooling to automatically learn important local features. PosPool \cite{pospool} and PCTP \cite{he2023pctp} improve the reduction by providing a position-adaptive pooling operation. Transformer-based methods \cite{pct,pointtransformerv1,pointtransformerv2,pointtransformerv3} use attention weight with softmax to sum neighboring points features. \cite{karambakhsh2022sparsevoxnet} and \cite{wang2025non} introduce additional spatial features as supplementary to enhance the feature representation of each point and facilitate the distinction of correlations. PointConvformer \cite{pointconvformer} and X-3D \cite{x-3d} utilize attention between neighboring points features to remove noise points during local aggregation in the convolution process. Removing the noise effect of irrelevant points can effectively improve the accuracy of extracting neighbor structures \cite{pointconvformer}.

Although existing works have begun to use supplementary features to strengthen the description of correlation, these methods mainly based on the calculation of single point-pairs correlation that may be disturbed by noise point. We propose an lightweight cross-stage spatial descriptor (LCSD), which can compare the difference between the individual point and the overall neighbor structure to obtain a more accurate correlation. Furthermore, we propose two operations: CDIP and CICS, to correct the distribution of feature in high-dimensional space during neighbor aggregation.

\subsection{Cross-stage Structure Embedding in Neighbor Aggregation}
Three-dimensional relative position coordinates are commonly used to encode spatial structure information \cite{pointnet,pointcnn,randla,convpoint,pointnext,pointvector,dehghanpour2024point}. Although 3D relative coordinates accurately characterize local spatial structures, their inability to encode inter-point correlations has prompted numerous studies to explicitly construct cross-stage structure encodings.

Some work \cite{large-scale-spg, spg-scalable, oa-cnn,patchcnn} aggregate neighboring points into small structural (e.g. superpoint or patch) that serve as fundamental computational units for direct network operations. These works usually have confusion problems in the details because the granularity of the division is relatively large. Some other work \cite{pointNN,pointhop,x-3d,zhu2025sgg} generate more accurate point-wise spatial information by encoding the neighbor distribution of each point, which can usually be used as an enhancement of the existing neighbor aggregation module to improve network performance.

However, the performance gains from point-wise cross-stage structure coding methods primarily stem from their direct augmentation of local structural features through geometric structural encoding (typically implemented via element-wise addition).  This necessitates higher encoding dimensions to ensure representational fidelity, incurring significant computational overhead. Moreover, such approaches constrain feature fusion to post-extraction stages, where independent processing often induces feature-space discrepancies and amplifies noise propagation.

We propose a neighbor aggregation method based on cross-stage correlation correction, which leverages high-dimensional feature correlations to dynamically adjust feature distributions during neighbor aggregation. Compared with the previous end-fusion methods, our method has lower dimensionality requirements for feature encoding and better robustness against noise point interference.

\section{Method}
\label{sec:method}
In Sec \ref{sec:preliminary}, we first introduce the neighbor aggregation process using local relative coordinates, as exemplified by the SA module. In Sec \ref{sec:analysis and improvement} we systematically analyzes feature noise problem in such neighbor aggregation process and propose the corresponding correction solution. In Sec \ref{sec:point-distribution set abstraction}, we introduced the specific implementation of the corresponding solution: PDSA module, which is based on LCSD to execute the CDIP and CISC correction operations. In Sec \ref{sec:network structure}, we introduce the network architecture in detail.

\subsection{Preliminary}
\label{sec:preliminary}
\begin{figure}[ht]
    \centering
    \vspace{0cm} 
    \subfigbottomskip=5pt 
    \setlength{\abovecaptionskip}{20pt} 

        \begin{minipage}[b]{1.0\linewidth}
            \centering
            \includegraphics[width=1.0\linewidth]{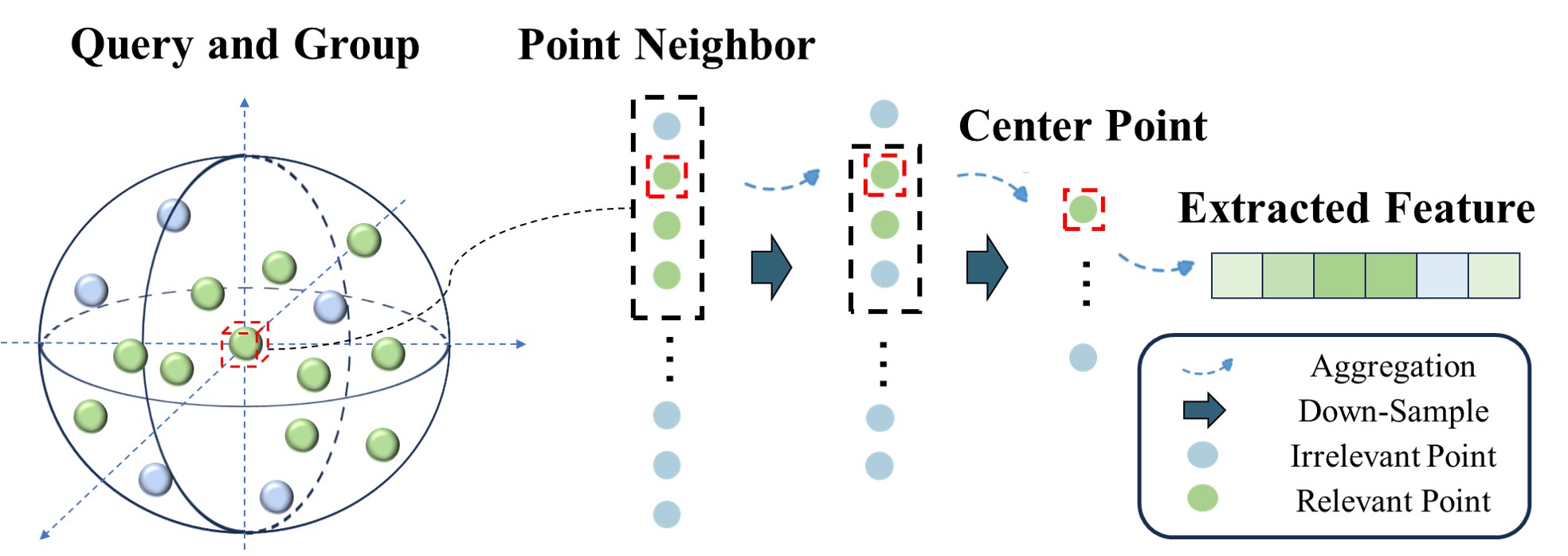}
        \end{minipage}

    \caption{Illustrations of hierarchical feature extraction process for point cloud analysis, with the SA module serving as the representative module. The green points represent relevant homogeneous points relative to the center point, while the blue points denote irrelevant points.}
    \label{fig1}
    \vspace{-8pt} 
\end{figure}

The Point Set Abstraction (SA) module is a neighbor aggregation module proposed by PointNet++ \cite{pointnet++}, which is widely referenced and adopted in subsequent series works \cite{pointnext,pointmeta,pointvector,pointnat} for extracting neighbor structure. These aggregation modules based on the relative three-dimensional coordinates of the point neighbor have similar extracting process, and this process is illustrated in Fig.\ref{fig1}: typically include down-sampling operation (Farthest Point Sampling or Random Sampling), group operation (K-NN or Ball Query), shared MLPs and reduction function (typically using MAX, AVG or SUM) to extract local structure from neighboring point relative coordinates.

We denote ${p}_{i} \in {R}^{1\times 3}$, ${f}_{i}\in {R}^{1\times c}$ as the $xyz$ coordinate and $c$-dimensional feature of point $i$ in current stage; denote $N$ as the set of current input point cloud and  ${N}_{i}$ as the neighbor of point $i$; denote ${p}_{i,j}$, ${f}_{i,j}$ as the coordinate and feature of neighboring point $j$ in point neighbor ${N}_{i}$, and $k$ is the number of points in the neighbor ${N}_{i}$. The feature aggregation can be written as follows:
\begin{align}
\label{eq:1}
{f}_{i}^{\prime }=\mathcal{R}\left \{ \mathcal{M}\left \{ {f}_{i,j},{p}_{i}-{p}_{i,j}\right \}\mid j\in {N}_{i}\right \}, i \in N
\end{align}
where $\mathcal{R}$ is the reduction function, $\mathcal{M}$ is the shared MLPs and ${f}_{i}^{\prime }$ is the aggregated feature. 
Let ${f}_{{N}_{ij}}$ denote embedded neighbor feature of ${N}_{i}$: 
\begin{align}
\label{eq:2}
{f}_{{N}_{ij}}=\mathcal{M}\left \{ {f}_{i,j},{p}_{i}-{p}_{i,j}\right \},j\in {N}_{i}
\end{align}
and the \cref{eq:1} can be written as:
\begin{align}
\label{eq:3}
{f}_{i}^{\prime }=\mathcal{R}\left \{ {f}_{{N}_{i}}\right \}
\end{align}
${f}_{i}^{\prime }\in {R}^{1\times c}$ is the next stage semantic feature. For such neighbor structure extraction process, \cref{eq:3} can be divided into two parts: 

1) Encoding the spatial structure of local neighbor: encoding the relative coordinates of shape $k \times 3$ into the neighbor feature matrix ${f}_{{N}_{i}}\in {R}^{k\times c}$.

2) Pooing the neighbor feature matrix: pooing $k$ encoded features in point neighbor ${N}_{i}$ on each channel with the reduction function, which is equivalent to aggregating each column of the neighbor feature matrix ${f}_{{N}_{i}}$.
\subsection{Analysis and Correction}
\label{sec:analysis and improvement}
We focus on analyzing the existing deficiencies in the ${f}_{i}^{\prime}$ calculation and propose the corresponding correction. We mainly consider corrections from two aspects: (1) the generalization problem caused by irrelevant point interference of neighbor feature matrix ${f}_{{N}_{i}}$ and (2) the feature separability affected by the description accuracy.

\subsubsection{Irrelevant Point in Neighbor Aggregation}
\label{subsec: Irrelevant Point in Neighbor Aggregation}
For the first problem, we focus on the neighbor feature matrix ${f}_{{N}_{i}}$, which is a $k\times c$ shape matrix obtained by MLP $\mathcal{M}$ extracting the relative coordinates $xyz$. Due to grouper's implementation mechanism, the sampled spatial neighbor point set is not always regular: K-NN sampling may result in an unfixed sampling range, while Ball Query sampling may result in refilling; the random sampling results will substitute interference factors such as irrelevant points into the extraction process of neighbor structure.

 \cite{rademacher} has proven that a smaller Gaussian complexity leads to better generalization, and \cite{filter,pointconvformer} demonstrate the upper bound of the Gaussian complexity of point cloud network using CNN architecture is determined by the following formula:
\begin{align}
\label{eq:gasussian bound}
\min \sideset{}{}{\max }_{{p}_{j}\in {N}_{i}}\sqrt{{\mathrm{\displaystyle\sum_{j=1}^{k}}}_{}{\left ( x\left ( {p}_{i}\right )-x\left ({p}_{j} \right )\right )}^{2}}
\end{align}
where $x\left ( {p}_{i}\right )$ denotes the the extracted feature of point ${p}_{i}$. Although this boundary is applied to a typical CNN network, it can also be implemented for shared MLP implemented with 1×1 convolution. The optimization goal of Eq.(\ref{eq:gasussian bound}) is to minimize the Euclidean distance difference between neighboring points features by selecting points that have high correlation belong to the same neighbor \cite{pointconvformer}. This means that each row in the ${f}_{{N}_{i}}$ matrix should be close in Euclidean distance. 

However, the feature extraction method in Eq.\ref{eq:1} is based on neighbor structure which is a $k \times c$ matrix, and this means that the point-to-neighbor  correlation cannot be derived from individual point coordinates. The inclusion of irrelevant points will induce deviations in the distribution of corresponding columns (i.e., feature channels) in matrix $f_{N_i}$.

Therefore, we need to introduce an \textbf{additional spatial feature $r_v$ that can describe the correlation between individual point and its neighbor} to correct the matrix ${f}_{N_i}$.
Further, in order to \textbf{reduce the distance deviation of the features between each row}, we propose a spatial encoding augmentation correction that leverages geometric differences between central points and their neighbors. 
 The improved ${f}_{{N}_{i}}$ calculation formula can be written as the following formula:
\begin{align}
\label{eq:fni improved}
{f}_{{N}_{ij}}=\mathcal{M}\left \{ {f}_{i,j},{p}_{i}-{p}_{i,j}\right \} + {Correct}({r}_{vij}-r_{vi}),j\in {N}_{i}
\end{align}
where ${r}_{vij}$ is the descriptor of neighboring point and ${r}_{vi}$ is the descriptor of the neighbor. This correction strategy improves the homogeneity of the neighbor feature matrix by enhancing the proportion of the feature distribution of the relevant points.

\subsubsection{Low Feature Separability Because of Hierarchical Gap}
\label{subsec: Low Feature Separability Because of Hierarchical Gap}
The second problem concerns separability between classes, which necessitates maximizing inter-distribution mean discrepancy while minimizing intra-class variance \cite{atheoretical}.

 Point clouds are typical manifold data in non-Euclidean space, where low-dimensional Euclidean coordinate representations may fail to characterize complex distribution patterns in high-dimensional embedding spaces.(e.g. Euclidean distances are very close, but geodesic distances are very far) \cite{x-3d}. Moreover, while relative local coordinates partially capture neighbor contexts, their descriptive capacity remains constrained by the limited receptive field of radial grouping operations. Therefore, these representational limitations potentially induce feature separability in high-dimensional latent spaces.

To solve the hierarchical gap between Euclidean coordinate and semantic feature, some approaches \cite{pointmeta,pointvector} explicitly the extraction neighbor structure by using different weight parameters for local coordinates and global features, and get has better experimental performance. Besides, some approaches \cite{pointnext,pointnat} which uses additional point-wise MLPs after SA module to reinforce the semantic feature ${f}_{i}$; this reinforcement of mapping to a high-dimensional space improves the high-dimensional distribution of feature and reduces information loss in aggregation \cite{mobilenetv2}. 

In summary, we attribute this to structural bias caused by the inability of local coordinates for accurately characterizing global distribution patterns. Therefore, we propose a correction operation for improving classes separability based on \textbf{additional cross-stage descriptor $r_g$, which can enhance the characterization of neighbor structural patterns} while simultaneously \textbf{transcending the receptive field constraints to model long-range contextual dependencies}. The correction can be written as the following improvement:

\begin{align}
\label{eq:fi improvement}
{f}_{i}^{\prime }=\mathcal{R}\left \{ {f}_{{N}_{i}}\right \} + Correct(r_{gi})
\end{align}
where $Correct(r_{gi})$ denotes the correction information based on additional cross-stage feature $r_{gi}$. The correction term $Correct(r_{gi})$ (including auxiliary global contextual information) enables direct distribution rectification through element-wise addition to the pooled feature $f_i^\prime$.

In summary, our proposed improvement can be written as following:
\begin{equation}
\label{eq:improved summary}
\begin{aligned}
{f}_{i}^\prime=\mathcal{R}\left \{\mathcal{M}\left \{ {f}_{i,j},{p}_{i}-{p}_{i,j}\right \} + {Correct}({r}_{vij})\right \} 
\\ + Correct(r_{gi}),j\in {N}_{i}
\end{aligned}
\end{equation}

\begin{figure*}[t]
    \centering
    \vspace{0cm} 
    \includegraphics[width=1\textwidth]{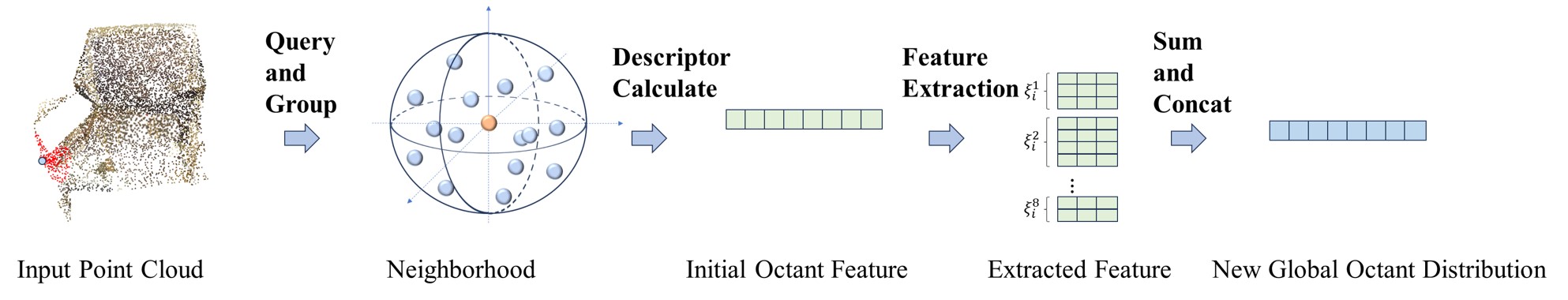}
    \caption{Illustration of the global octant distribution feature calculation. The figure shows four calculation steps in sequence: 1) Query and group were used to obtain the neighboring points from the input point cloud; 2) Calculate the octant distribution description in the neighbor; 3) Extract the octant feature of the neighboring points according to the octant distribution; 4) Sum and concat the extracted features in the octant order.}
    \label{fig:octant feature}
    \vspace{-10pt} 
\end{figure*}
\subsection{Point-Distribution Set Abstraction}
\label{sec:point-distribution set abstraction}
In this subsection, we will introduce the Point-Distribution Set Abstraction which is the implementation of the improved method in Sec \ref{sec:analysis and improvement}. We propose the LCSD to distinguish the correaltion of individual point, and design the corresponding correction operations (CDIP and CICS) to solve the problems anaylized in SubSec \ref{subsec: Irrelevant Point in Neighbor Aggregation} and SubSec \ref{subsec: Low Feature Separability Because of Hierarchical Gap}.
\subsubsection{Lightweight Cross-stage Spatial Distribution Descriptor}
As discussed in SubSec \ref{subsec: Irrelevant Point in Neighbor Aggregation} and \ref{subsec: Low Feature Separability Because of Hierarchical Gap}, we need additional descriptor $r_v$ and $r_g$ to distinguish irrelevant points for reducing the distance deviation in $f_{N_i}$ and enhance the characterization of neighbor structural
patterns for improving distributional separability of $f_i^\prime$. We notice that there are numerous work \cite{pointNN,pointhop,x-3d} based on spatial structure embedding can perform point cloud analysis tasks with non/less learning parameters, which means that the spatial structure information of point cloud can provide guidance for semantic distribution.

Inspired by this observation, we introduce \textbf{a lightweight cross-stage structure descriptor ($ACSD$) $d$ to unify $r_v$ and $r_g$ solving the above both problems simultaneously}: (1) discriminating individual point correlation through spatial distribution similarity metrics and (2) enhancing high-dimensional feature separability via augmented distributional characterization.

Previous work PointHop\cite{pointhop} and X-3D\cite{x-3d} has encoded cross-stage structure based on octant centroid. However, this encoding scheme exhibits two critical limitations: susceptibility to noise from edge-irrelevant points and parameter inefficiency. To address these issues, we enhance this approach by incorporating a distance-aware relative weighting mechanism that adaptively weight octant distributions during cross-stage aggregation. This refinement achieves dual objectives: (1) improving spatial correlation preservation during cross-stage aggregation, and (2) reducing  parameter cost through a more concise encoding method. 

The calculation process of this descriptor is as follows:
\begin{align}
\label{eq:dis concat}
{d}_{i}^{l}=concat<{O}_{i}^{l,1},{O}_{i}^{l,2},...,{O}_{i}^{l,8} >
\end{align}
\begin{align}
\label{eq:dis_o calculate}
{O}_{i}^{l,o}=\displaystyle\sum_{j=1}^{k}{t}_{ij}^{l-1,o}{r}_{ij}{a}_{ij}^{l-1}, ~~~l=0,1,...,~~ o=1,2,...,8
\end{align}
where ${d}_{i}^{l}$ denotes the lightweight structure descriptor of point $i$ in stage $l$, ${O}_{i}^{l,o}$ denotes the octant distribution feature, ${a}_{ij}^{l-1}$ is the dimensionally reduced descriptor of ${d}_{i}^{l-1}$ extracted from the linear layer, ${t}_{ij}^{o}$ represents whether the neighboring point ${p}_{ij}$ belongs to the $o$-th octant:
\begin{align}
\label{eq:tij}
{t}_{ij}^{o}=\left\{\begin{matrix}1,~~ {p}_{ij}\in {\xi }_{i}^{o}
\\ 
0,~~{p}_{ij}\notin  {\xi }_{i}^{o}
\end{matrix}\right.
\end{align}

The ${r}_{ij}$ is distant relative weight which decreases from $1$ to $0$ as the distance increases, and the addition of this weight improves the accuracy of the aggregation results. The process is shown in Fig.\ref{fig:octant feature}.
\begin{align}
\label{eq:rij}
{r}_{ijj}=\left\{\begin{matrix}1-{x}_{ij}/radius,~~ &{p}_{ij}\ne {p}_{i}
\\ 
0,~~~~~&{p}_{ij} = {p}_{i}
\end{matrix}\right.
\end{align}

\subsubsection{Correction for De-noising Irrelevant Points}
To achieve the correction in SubSec \ref{subsec: Irrelevant Point in Neighbor Aggregation}, we quantify the correlation degree of the neighboring points via lightweight cross-stage descriptor (LCSD) $d_i$ while minimizing the Euclidean distance variance of the neighbor feature matrix $f_{Ni}$ for better generalization.

To distinguish irrelevant point in neighbor aggregation, \cite{pointconvformer} compares the semantic feature difference between neighboring points, and reducing the $f_{N_i}$ distribution boundary through adaptive feature weighting; \cite{pointtransformerv1,pointtransformerv2,pct}  compute attention weights to focus on the relevant points, \cite{yujie2024s2anet} compare the point-wise correlation based on the similarity in the frequency domain and select the sampling points accordingly.

However, existing approaches predominantly focus on pairwise correlations among neighboring points while neglecting structural dependencies between individual points and their neighbor, potentially amplifying interference from irrelevant points.

Consequently, we establish the correlation between local points and their structural neighbor relationships through the LCSD $d$ as the denoising evaluation criterion, rather than pairwise point correlation.
Eq.(\ref{eq:dis_o calculate}) and Eq.(\ref{eq:dis concat}) shows the aggregation of distribution feature ${d}_{i}^{l}$, therefore, the next-stage distribution feature ${d}_{i}^{l+1}$ can describe the overall spatial structure of the neighbor, which means ${d}_{i}^{l+1}$ is equivalent to the neighbor structure descriptor ${d}_{{N}_{i}}$. We can calculate the correlation of neighboring point by comparing ${d}_{ij}^{l}$ and ${d}_{i}^{l+1}$.

To address this, we minimize semantic feature discrepancies within local neighbor through a cross-attention-inspired mechanism that simultaneously captures individual point correlations and structural differences, finally generating adaptive attention weights and corrective structural coding to enhance the homogeneity of the overall neighbor structure.
For better local geometric invariance, we also use the distribution difference between overall neighbor and each neighboring point as the structural calculation criterion:
\begin{equation}
\label{eq:wre}
{W}_{rej} =SoftMax\left \{ {M}_{W}\left \{ {L}_{q}({d}_{i}^{l+1})-{L}_{k}({d}_{ij}^{l}) \right\} \right\}
\end{equation}

\begin{align}
\label{eq:st}
{V}_{stj} = {M}_{v}\left \{ {L}_{q}({d}_{i}^{l+1})-{L}_{k}({d}_{ij}^{l}) \right\} ,j \in N_i
\end{align}
where ${L}_{q}$ and ${L}_{k}$ are the linear layer used for embedding queries and key values, respectively; ${M}_{w}$ and ${M}_{v}$ are the MLPs used for calculating attention-weight and embedding local structure. To summarize the above, improved $f_{Ni}$ calculation can be written as follow:
\begin{align}
\label{eq:improved fni implementation}
{f}_{{N}_{ij}}=\mathcal{M}\left \{ {f}_{i,j},{p}_{i}-{p}_{i,j}\right \}+W_{rej}\times V_{stj},~~j \in N_i
\end{align}

The specific network structure design is shown in the Fig.\ref{fig:framework}, where the MLP layer is implemented by $1 \times 1$ convolution.

\subsubsection{Correction for Improving Classes Separability}
As we discussed in SubSec \ref{subsec: Low Feature Separability Because of Hierarchical Gap}, due to the limitation of grouper sampling radius and the characteristics of point cloud non-Euclidean spatial manifold data, there may be distribution confusion between semantic features encoded by local relative three-dimensional coordinates.

We solve this problem by introducing a global self-attention mechanism based on lightweight cross-stage spatial descriptor (LCSD) $d$ to enhance the separability between features. Because of the hierarchical consistency between LCSD $d_i$ and semantic feature $f_i$, along with the information-rich representations from local neighbor, the simplified feature $d_i$ effectively achieves accurate correction.

Intuitively, for correction of classes separability can be designed directly as a global self-attention based on $d_i$; the eq.\ref{eq:fi improvement} can be rewritten as follow:
\begin{align}
\label{eq:fi sat correction}
{f}_{i}^{\prime }=\mathcal{R}\left \{ {f}_{{N}_{i}}\right \} + SAT(d_{i})
\end{align}
where $SAT$ denotes the self-attention mechanism. Eq.\ref{eq:fi sat correction} reinforces the divisibility of the feature distribution by adopting self-attention to the spatial distribution of each point. However, such an approach would introduce an unacceptable additional computational overhead when dealing with large scene point cloud due to the $QKV$ matrix multiplication of the attention mechanism.
\begin{figure}[ht]
    \centering
    \vspace{0cm} 
    \subfigure[]{
        \begin{minipage}[b]{.475\linewidth}
            \centering
            \label{fig1:subb}
            \includegraphics[width=1.0\linewidth]{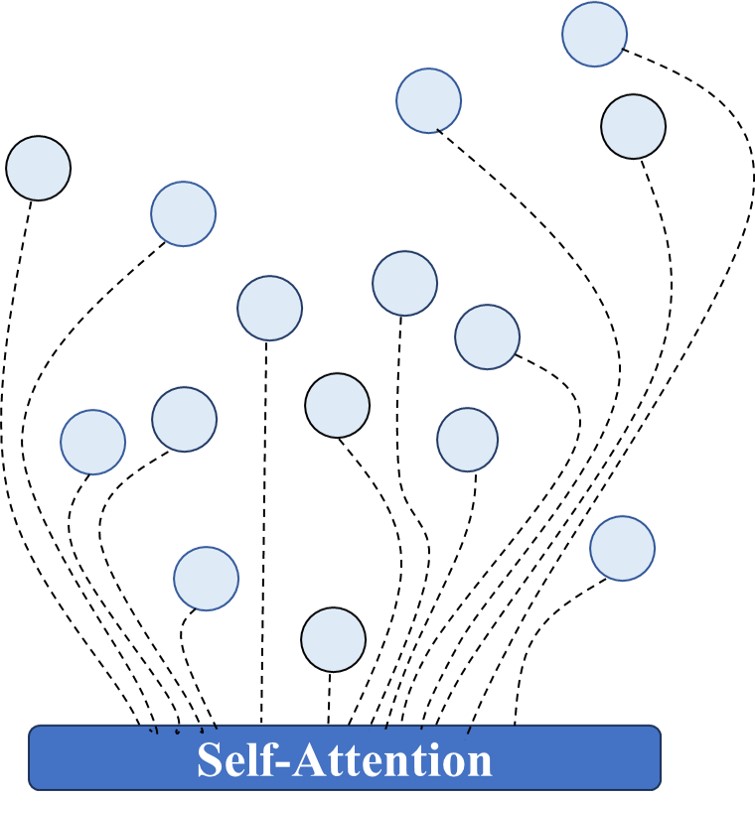}
        \end{minipage}
    }
    \subfigure[]{
        \begin{minipage}[b]{.45\linewidth}
            \centering
            \label{fig1:subc}
            \includegraphics[width=1.0\linewidth]{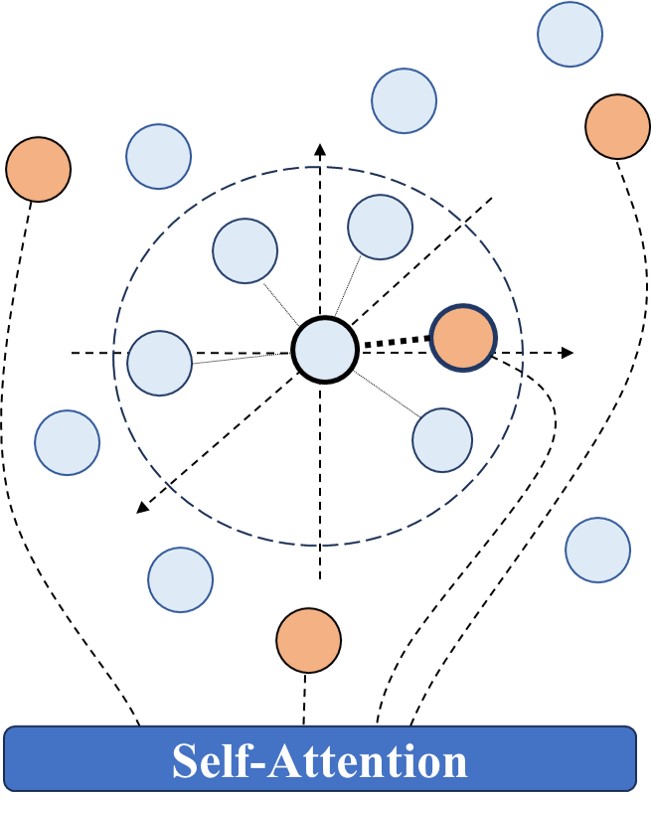}
        \end{minipage}
    }
    \caption{Illustration of the matching mechanism of global self-attention based on Key Point. Blue points represent the ordinary points that do not participate in the operation and the orange points denote the key point which involved in the operation of the global attention mechanism. (a) The illustration of native global self-attention implementation. (b) The illustration of global self-attention implementation based on key point selection.)}
    \label{fig:sat}
    \vspace{-8pt} 
\end{figure}

The core objective revolves around maximizing inter-class separation while minimizing intra-class variance in high-dimensional embedding spaces. Guided by this principle, our methodology strategically selects representative points for global attention computation in large-scale point cloud scenarios, eliminating the need for exhaustive point-wise participation while maintaining effective class discriminability.

In details, attention weights computed via Eq.(\ref{eq:wre}) are aggregated per point, with key points subsequently selected as representative points through ranking-based thresholding. The key point selection process involves computing statistical aggregates of attention weights across neighboring points, followed by globally ranked selection of optimal representatives.  This strategy enhances intra-neighbor homogeneity through geometrically consistent sampling from the most discriminative feature subspaces. The specific visual analysis on the representativeness of key points is in Sec.\ref{sec:ablation}.
Therefore, in the large scene point cloud, we only calculate the key points for correction information by global self-attention, and then copy the correction information of the key points to the rest of the points in the neighbor. This process is shown in Fig.\ref{fig:sat}.


\begin{figure*}[ht]
    \centering
    \vspace{0cm} 
    \includegraphics[width=1\textwidth]{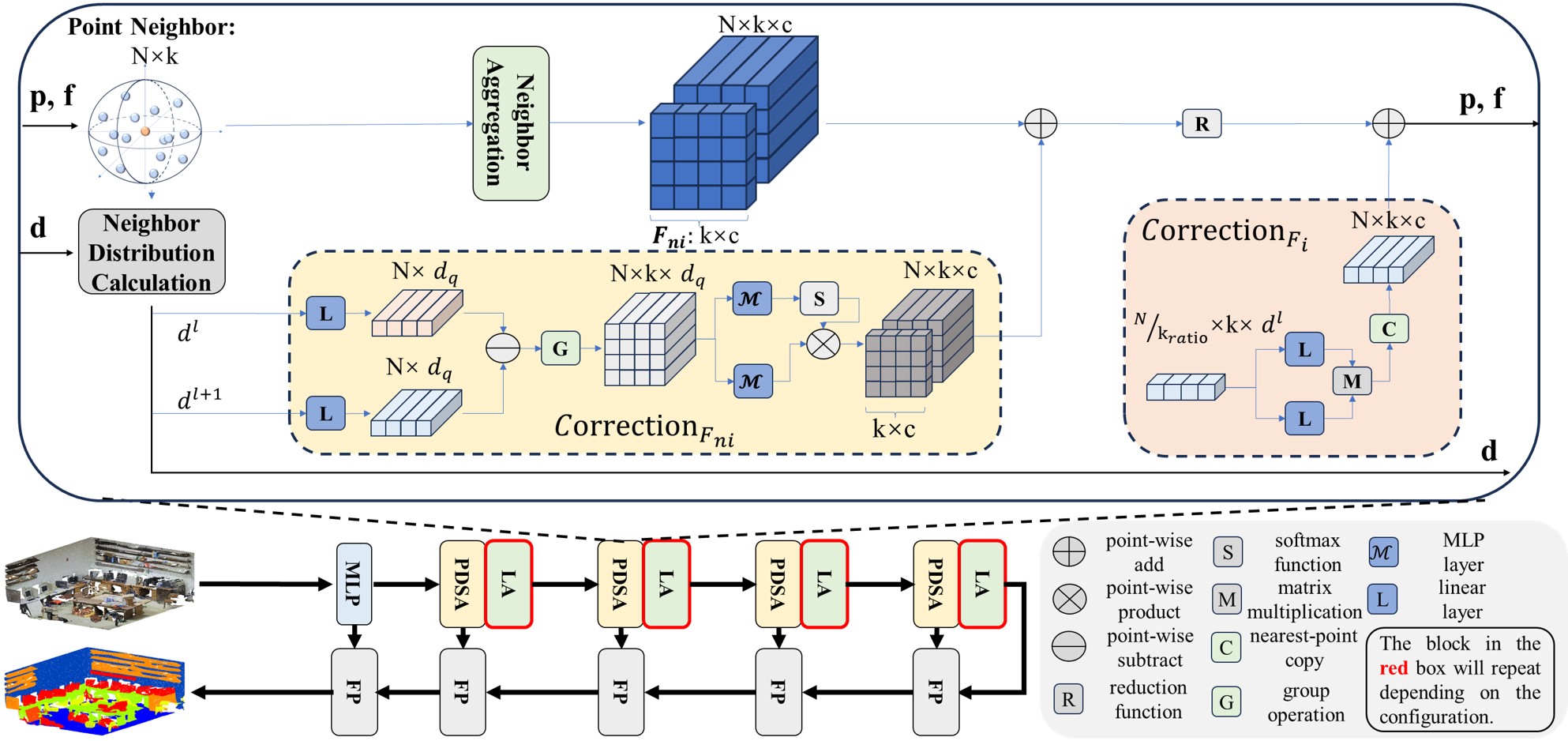}
    \caption{Illustration of the network architecture. FP is the same Feature Propagation block as PointNet++\cite{pointnet++} and PointNext\cite{pointnext}. LA is the Local aggregation block and has two versions: PDSA duplicate version for PointDistribution-base and VPSA\cite{pointvector} version for PointDistribution-vp. The overall input and output of PDSA are represented by $p,f,d$, which respectively represent the three-dimensional coordinates, semantic features and octant distribution features of the point cloud.}
    \label{fig:framework}
    \vspace{-10pt} 
\end{figure*}
\subsection{Network Architecture}
\label{sec:network structure}
Based on the proposed PDSA, we design the network architecture named as PointDistribution (PD) as shown in Fig.\ref{fig2}, which is on the basis of the PointNet modern framework OpenPoint \cite{pointnext}.

For semantic segmentation tasks, the network structure adopts a U-net like architecture includes encoder and decoder. In the feature coding stage, its modules can be divided into SA block and Local Aggregation (LA) block: SA block contains both down-sampling and neighbor aggregation, and LA block only contains neighbor aggregation for refining local features. We use PDSA to replace Point Set Abstraction module in SA block, while different local aggregation modules can be selected for LA block. 

For the selection of LA block, it can be a simple repetition of SA module or other modern aggregation module based on pointNet++ \cite{pointnext,pointmeta,pointvector}
, and PointVector \cite{pointvector} is one of them, which uses vector representation to enhance feature expression capability in local feature aggregation. Therefore, we use the VPSA module proposed by PointVector to conduct comparative experiments. For the classification task, we only use the encoder part of the network architecture.

The network can be configured in different scales. Let $L$ denote the number of LA blocks, and let $C$ denote the dimensions of the channel. We denote the PD-base as the version where PDSA module is chosen for the LA block, and denote the PD-vp as the version that uses VPSA module in LA block. The configuration is summarized as follows:
\begin{itemize}
\item PD-base-S: $C$ = 32, $L$ = 0
\item PD-base-XL: $C$ = 64, $L$ = [3, 6, 3, 3]
\item PD-vp-XL: $C$ = 64, $L$ = [3, 6, 3, 3]
\end{itemize}
where the $L$ of PD-base-S is $0$, which means that no LA block used in net work.
\section{Experiments}
\label{sec:Experiment}
To demonstrate the effectiveness of our approach, we conduct experiments on three standard benchmarks for semantic segmentation and classification tasks, i.e., S3DIS \cite{s3dis} for segmentation, ScanObjectNN \cite{scanobjectnn} and ModelNet40 \cite{modelnet40} for classification. We evaluated all our model on a Nvidia Geforce RTX 4090 24-GB GPU with a 12-core Intel Xeon(R) Platinum 8352V CPU @2.10Ghz. We trained our model using CrossEntropy loss with label smoothing optimized by AdamW optimizer for all tasks, and the initial dimension of additional cross-stage spatial descriptor $d_{i}^{0}$ is set to 8, and each reduction is set to $3^{l}$ in layer $l$. 

For evaluation methods, we refer to previous work, using mean intersection over union (mIoU) and overall accuracy (OA) for semantic segmentation task and OA and mean accuracy (mAcc) for classification task.
\begin{align}
\text{miou} &= \frac{1}{n} \sum_{i=1}^{n} \left( \frac{TP_{i}}{FN_{i} +  FP_{i} + TP_{i}} \right)
\end{align}
\begin{align}
\text{OA} &= \frac{\sum_{i=1}^{n} TP_{i}}{n} 
\end{align}
\begin{align}
\text{mAcc} &= \frac{1}{n} \sum_{i=1}^{n} \left( \frac{TP_{i}}{TP_{i} + FN_{i}} \right) 
\end{align}
where TP denotes the true positive samples, FP denotes the false positive samples, FN denotes the false negative samples, $i$ is the $i$-th indicator in the total number $n$ of semantic classes.
\subsection{Semantic Segmentation in S3DIS}
\textbf{S3DIS} \cite{s3dis} (Stanford Large-Scale 3D Indoor Spaces) is a challenging benchmark of indoor scene point cloud semantic segmentation which was reconstructed from RGB-D images captured by camera equipped with structured light sensors. Each point in this dataset has been meticulously annotated into 13 distinct semantic classes.

For model training, we follow the previous work \cite{pointnext,pointvector} and adopt the same configuration: the input point cloud is downsampled with a voxel size of 0.4 m and the number of input is fixed at 24000 points per batch; the initial learning rate is set to 0.01, and the weight decay is set to $10^{-4}$; the data enhancements employed are also same as the previous work. PD-base sets 8 batches per epoch and PD-vp sets 6 batches per epoch, with 100 epochs of training in total. 
We evaluate our model using the best model of validation set to test the entire scene of S3DIS Area5, and the results are shown in Tab.\ref{tab:s3dis}.
\begin{table*}[ht]
\setlength{\tabcolsep}{4pt}
    \centering
    \small
    \caption{Semantic Segmentation Results on The S3DIS Benchmark.}
    \begin{tabular}{lcccccccccccccccc}
        \toprule
         Method & \rotatebox{90}{params(M)}  & \rotatebox{90}{OA(\%)} & \rotatebox{90}{mIou(\%)} & \rotatebox{90}{ceiling} & \rotatebox{90}{floor} & \rotatebox{90}{wall} & \rotatebox{90}{beam} & \rotatebox{90}{column} & \rotatebox{90}{window} & \rotatebox{90}{door} & \rotatebox{90}{table} & \rotatebox{90}{chair} & \rotatebox{90}{sofa} & \rotatebox{90}{bookcase} & \rotatebox{90}{board} & \rotatebox{90}{clutter} \\
        \midrule
        PointNet\cite{pointnet} &3.6  & - & 41.1 & 88.8 & 97.3 & 69.8 & 0.1 & 3.9 & 46.3 & 10.8 & 59.0 & 52.6 & 5.9 & 40.3 & 26.4 & 33.2 \\
        
        PointNet++\cite{pointnet++}&1.0 & 83.0 & 53.5 & - & - & - & - & - & - & - & - & - & - & - & - & - \\
        
        PointNet++(OP)\cite{pointnext} & 1.0  & 88.3 & 63.6 & 93.6 & 98.4 & 81.3 & 0.0 & 22.3 & 53.0 & 70.4 & 80.5 & 87.0 & 55.1 & 70.8 & 61.8 & 52.8 \\
        
        KPConv\cite{kpconv} & 15.0  & - & 67.1 & 92.8 & 97.3 & 82.4 & 0.0 & 23.9 & 58.0 & 69.0 & 81.5 & 91.0 & 75.4 & 75.3 & 66.7 & 58.9 \\
        PAConv\cite{paconv} & -  & - & 66.6 & 94.6 & \textbf{98.6} & 82.4 & 0.0 & 26.4 & 58.0 & 60.0 & \textbf{89.7} & 80.4 & 74.3 & 69.8 & 73.5 & 57.7 \\
        PT\cite{pointtransformerv1} & 7.8  & 90.8 & 70.4 & 94.0 & 98.5 & \textbf{86.3} & 0.0 & 38.0 & \textbf{63.4} & 74.3 & \underline{89.1} & 82.4 & 74.3 & 80.2 & 76.0 & 59.3 \\
        PointMetaBase\cite{pointmeta} & 15.3 & 90.9 & 71.1 & - & - & - & - & - & - & - & - & - & - & - & - & - \\
        SegPoint\cite{segpoint} & - & - & 72.4 & - & - & - & - & - & - & - & - & - & - & - & - & - \\
        
        ConDaFormer\cite{condaformer} & -  & 91.6 & 72.6 & - & - & - & - & - & - & - & - & - & - & - & - & - \\
        
        SAT\cite{zhou2023sat} & - &  - & 72.6 & 93.6 & 98.5 & 87.2 & 0.0 & \underline{49.3} & 61.1 & 73.6 & 83.7 & 91.8 & 81.7 & 77.9 & 82.3 & \underline{63.4} \\ 
        
        PointNAT\cite{pointnat} & 24.9 & 91.5 & 72.8  & - & - & - & - & - & - & - & - & - & - & - & - & - \\
        
        PointHR\cite{qiu2023pointhr} & - & \underline{91.8} & 73.2  & 94.0 & 98.5 & 87.5 & 0.0 & \textbf{53.7} & \underline{62.9} & 80.2 & 84.2 & 92.5 & 75.4 & 76.5 & 84.8 & 61.8 \\
        
        SPG\cite{han2024subspace} & - & \textbf{91.9} & \underline{73.3}  & - & - & - & - & - & - & - & - & - & - & - & - & - \\
        
        \midrule

        PointNeXt-XL\cite{pointnext} & 41.6  & 90.6 & 70.5 & 94.2 & 98.5 & 84.4 & 0.0 & 37.7 & 59.3 & 74.0 & 83.1 & 91.6 & 77.4 & \underline{77.2} & 78.8 & 60.6 \\
        \rowcolor{gray!20}
        \textbf{PD-base-XL(ours)} & 12.5  & 91.0 & 71.2 & 94.5 & \textbf{98.6} & \underline{86.0} &  0.0 &   47.3 &  58.4 & 73.4 & 83.3 & 91.7 & 76.4 & 76.1 & 80.0 & 60.3 \\
        
        \midrule
         PointVector-XL\cite{pointvector} & 24.1  & 91.0 & 72.3 & \underline{95.1} & \textbf{98.6} & 85.1 & 0.0 & 41.4 & 60.8 & \underline{76.7} & 84.4 & \underline{92.1} & \underline{82.0} & \underline{77.2} & \textbf{85.1} & 61.4 \\
         \rowcolor{gray!20}
        \textbf{PD-vp-XL(ours)} & 28.8  & \underline{91.8} & \textbf{73.4} & \textbf{95.7} & \textbf{98.6} & 85.7 &  0.0 &  43.7 & 57.9 & \textbf{78.4} & 85.3 & \textbf{92.6} & \textbf{88.0} &  \textbf{78.9} & \underline{83.5} & \textbf{65.9} \\
        \bottomrule   
    \end{tabular}
    \parbox{\textwidth}{PointNet++(OP) is the version of PointNet++\cite{pointnet++} deployed under the openpoint\cite{pointnext} code framework, which has been strengthened by the training strategy.}
    \label{tab:s3dis}
\end{table*}

PointNet++ (OP) denotes the version of PointNet++ \cite{pointnet++} which has been strengthened by the training strategy proposed in \cite{pointnext}. PointNext \cite{pointnext} is the modern version of PointNet++ \cite{pointnet++} which strength the training strategy and introduce the local aggregation block, PointVector \cite{pointvector} and our PD-base both adopt it as the baseline.

The experimental results show that our PD-base-XL model which is completely composed of PDSA module outperforms the PointNet++(OP) by \textbf{7.6\% mIoU} and \textbf{2.7\% OA} respectively, and outperform the PointNext \cite{pointnext} by \textbf{0.7\% mIoU} and \textbf{0.4\% OA} respectively under the case of \textbf{significantly reducing 70\% parameters}. PD-vp-XL outperforms the baseline PointVector \cite{pointvector} by \textbf{0.8 \% mIoU} and \textbf{1.1\% OA} respectively. 

By comparing the experiments of PD-base and PD-vp, adopting PDSA module improves the result significantly compared with the corresponding baseline; this verifies the effectiveness of our proposed neighbor distribution correction aggregation to improve the generalization of aggregated features.

\begin{figure*}[ht]
    \centering
    \vspace{0cm} 
    \subfigbottomskip=5pt 
    \setlength{\abovecaptionskip}{20pt} 

        \begin{minipage}[b]{1.0\linewidth}
            \centering
            \includegraphics[width=1.0\linewidth]{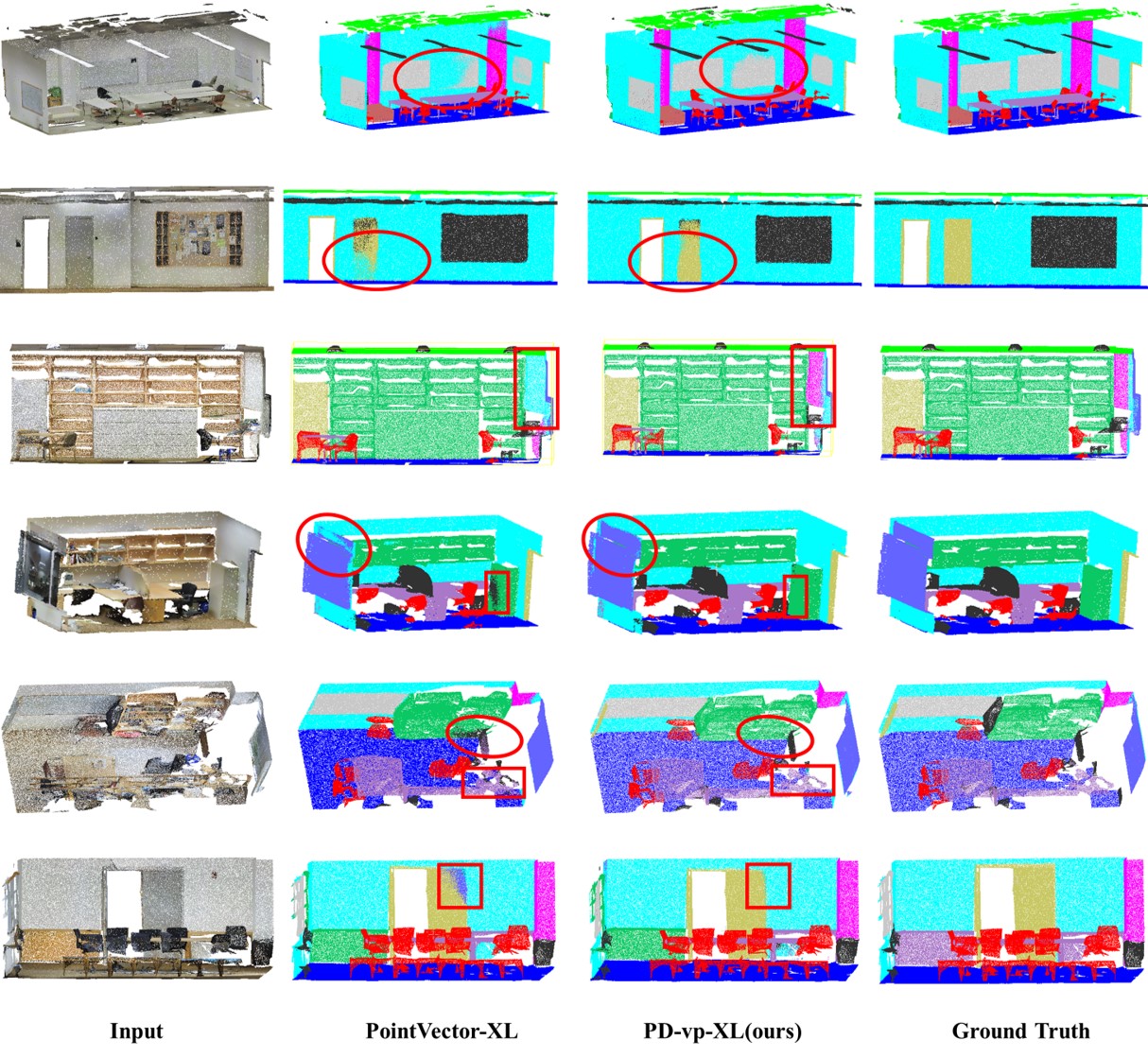}
        \end{minipage}
    \caption{Comparison of the experimental results of PointVector-XL and PD-vp-XL on S3DIS area5. The red circles represent the segmentation defects with unclear boundaries. The red rectangular box represents the segmentation defect caused by classification errors.}
    \label{fig:s3dus reusult}
    \vspace{-8pt} 
\end{figure*}

The results of experiments on S3DIS area5 are shown in Fig.\ref{fig:s3dus reusult}. The visual segmentation results demonstrate that PD-VP achieves sharper boundary delineation compared to PointVector. Segmentation visualization results demonstrate that our proposed PD-vp-XL (ours) achieves two critical improvements over PointVector \cite{pointvector}: (1) Significantly reduces boundary ambiguity at object interfaces (e.g. the segmentation result of the blackboard and the door in the first and second rows of Fig.\ref{fig:s3dus reusult}). (2)  Enhances the classification accuracy for objects with similar shapes (e.g. the segmentation result of the column and the door in the third and last rows of Fig.\ref{fig:s3dus reusult}).

This improvement indicates that the distribution correction mechanism in the PDSA module effectively mitigates noise interference from irrelevant points at object junctions. Meanwhile, since PDSA breaks through the limitation of the grouper sensory field by using global attention, it can also be observed in the Fig.\ref{fig:s3dus reusult} that PD-VP is more accurate in the classification of structurally similar and adjacent objects, which confirms our improvement in classes separability.

\subsection{Classification in ScanObjectNN and ModelNet40}
For classification task, we evaluate our model on the two famous benchmarks: ScanObjecNN \cite{scanobjectnn} and ModelNet40 \cite{modelnet40}, and we adopt a smaller network structure for classification following previous work \cite{pointnext}, i.e., PD-S, which is entirely composed of PDSA module and therefore no longer distinguishes extra versions.

\textbf{ScanObjectNN} \cite{scanobjectnn} contains approximately 15,000 real scanned objects that are categorized into 15 classes with 2,902 unique object instances. This is a significant challenge for existing point cloud analysis methods because of the occlusions and noise. We experiment on PB$\_$T50$\_$RS, the hardest and most commonly used variant of this dataset \cite{pointnext}. We train our model in 0.002 learning rate with a weight decay of 0.05 for 250 epochs, and set the number of input points to 1024. For the PD-s on the ScanObjectNN, the number of batch size is set to 32. 

\textbf{ModelNet40} is a famous and commonly used 3D object classification dataset, which has 40 object categories, each of which contains 100 unique CAD models. The train configuration is similar as ScanOjectNN, and the total number of training epochs is set to 600. The batch size is also set to 32.
\begin{table}[ht]
\setlength{\tabcolsep}{0.8pt}
    \centering
    \small
    \caption{Classification Results On ScanObjectNN And ModelNet40.}
    \begin{tabular}{lccccc}
        \toprule
        Method & \multicolumn{2}{c}{ScanObjectNN} & \multicolumn{2}{c}{ModelNet40} & Params.  \\
        \cmidrule(lr){2-3} \cmidrule(lr){4-5}
         & OA(\%) & mAcc(\%) & OA(\%) & mAcc(\%) & M  \\
        \midrule
        PointNet\cite{pointnet}& 68.2 & 63.4 & 89.2 & 86.2 & 3.5 \\
        PointNet++ [\cite{pointnext}] & 77.9 & 75.4 & 91.9 & - & 1.5  \\
        PointCNN\cite{pointcnn} & 78.5 & 75.1 & 92.2 & 88.1 & 0.6  \\
        KPConv\cite{kpconv} & - & - & 92.9 & - & 14.3  \\
        PCT\cite{pct} & - & - & \textbf{93.2} & - & 2.9 \\
        DGCNN(ACN)\cite{putra2025adacrossnet}  &  82.1 & - & \underline{93.1} & - & -\\
        PointStack\cite{pointstack} & \underline{86.9} & \underline{85.8} & \textbf{93.2} & 89.6 & -  \\
        PCM-Tiny\cite{pcm} & \underline{86.9} & 85.0 & \underline{93.1} & \textbf{90.6} & 6.9 \\
        \midrule
        \textbf{PD-S (ours)} & \textbf{87.3} & \textbf{85.5} & \underline{93.1} & \underline{90.4} & 2.5  \\
        \bottomrule
    \end{tabular}
    \label{tab:classification}
\end{table}

As the result shown in Tab.\ref{tab:classification}, our work achieves a large performance improvement compared with PoinNet++ \cite{pointnet++} with a small parameter cost, with improvements of \textbf{19.1\% OA, 22.1\% mAcc and 3.9\% OA, 4.2\% mAcc} on ScanObjectNN and ModelNet40 datasets respectively. Its performance is comparable to recent works. Because we adopt the approach of combining global sequence modeling with local feature extraction, it is also more efficient in terms of parameters than common serialization methods \cite{pct,pcm}.

\subsection{Ablation Study}
\label{sec:ablation}
In order to further verify the effectiveness of PDSA module, we choose the PointDistribution-base for ablation study, and the test data set is S3DIS-Aera5 \cite{s3dis} for semantic segmentation. For fair comparison, we did not change the training parameters.
\paragraph{Ablation of PDSA Module}
\label{par:ablation of PDSA}
We abstract the module into three key operations: (1) the correction operation CDIP for denoising irrelevant point in point neighbor aggregation, (2) the distant wight $D_w$ for refining LCSD description and (3) the correction operation CICS for improving classes separability through global self-attention mechanism. To analyze the effectiveness of each operation in the PDSA module, we designed a series of ablation experiments as shown in Tab.\ref{tab:ablation of PDSA module}.

The ablation results are shown in \cref{tab:ablation of PDSA module}. We can intuitively see that after added supplementary CDIP operation to baseline, mIou has been significantly improved by 5.1\% due to reducing the influence of irrelevant point on neighbor feature matrix ${f}_{{N}_{i}}$. In addition, with the addition of the distant weight, the mIou increases by 1.1\% compared with none $D_w$, which means that our designed distant weight enhances the descriptive ability of LCSD through refining the correlation during cross-stage aggregation. Furthermore, the operation CICS has increased mIou by 1.1\%, due to increase of receptive field for improving the classes separability. Finally, our proposed PD-base achieves the best performance after aggregating all operations, which validates the effectiveness of each operation. 

\begin{table}[h]
\setlength{\tabcolsep}{4pt}
  \centering
  \small
    \caption{Ablation Results of PDSA module.}
  \begin{tabular}{lcc}
    \toprule
Ablate & mIoU  & $\triangle$  \\
    \midrule
baseline(PointNet++)& 63.6 &  - \\
\hline
$\rightarrow CDIP$ & 68.7 &  \textbf{+5.1} \\
$\rightarrow CDIP+D_w$ & 69.8 &  \textbf{+1.1} \\
$\rightarrow CDIP+D_w+CICS(PD-base)$ & 71.2 & \textbf{+1.4}  \\

    \bottomrule
  \end{tabular}
  \label{tab:ablation of PDSA module}
\end{table}
\paragraph{Method of De-nosing Irrelevant Point}
\label{par:ablation of metod of denosing}
As discussed in\cref{sec:analysis and improvement}, we increase the generalization of network by calculating correlation between neighboring points and adding correction CDIP to neighbor features $k \times c$ matrix $f_{N_I}$ for de-noising the irrelevant points as shown in \cref{eq:fni improved} and \cref{eq:improved fni implementation}, rather than weighting individual points. To verify the effectiveness of this de-noising method, we design the following experiments based on PD-base-XL model to compare the effect of correcting the neighborhood feature matrix and weighting the single neighboring point. 
\begin{quotation}
$A1$: Based on ${d}_{ij}^l$ and ${d}_{i}^{l+1}$, calculating the correlation weight ${w}_{ij}$ and distribution structure ${v}_{ij}$ for neighboring points; Correcting the neighborhood features ${f_{{N}_{i}}}$ by point-wise multiplication: ${w}_{ij} \times ({v}_{ij}+\mathcal{M}\left\{ {p}_{ij} \right\} )$   

$A2$: Allocating weight for each ${d}_{ij}$ based attention mechanism, calculating the aggregated values ${v}_{i}^{\prime}$ by using $linear$ and $SoftMax$, and adding it to ${f}_{i}^{\prime}$ for correction.: $f_i^\prime=\mathcal{R}\left \{ {f}_{{N}_{i}}\right \}+\mathcal{R}\left \{ w_ij\times v_ij\right \}$ 

$A3$: The PD-base-XL which is entirely composed of PDSA.
\end{quotation}

\begin{figure*}[ht]
    \centering
    \vspace{0cm} 
    \subfigure[]{
        \begin{minipage}[b]{.49\linewidth}
            \centering
            \label{fig:vis sub1}
            \includegraphics[width=1.0\linewidth]{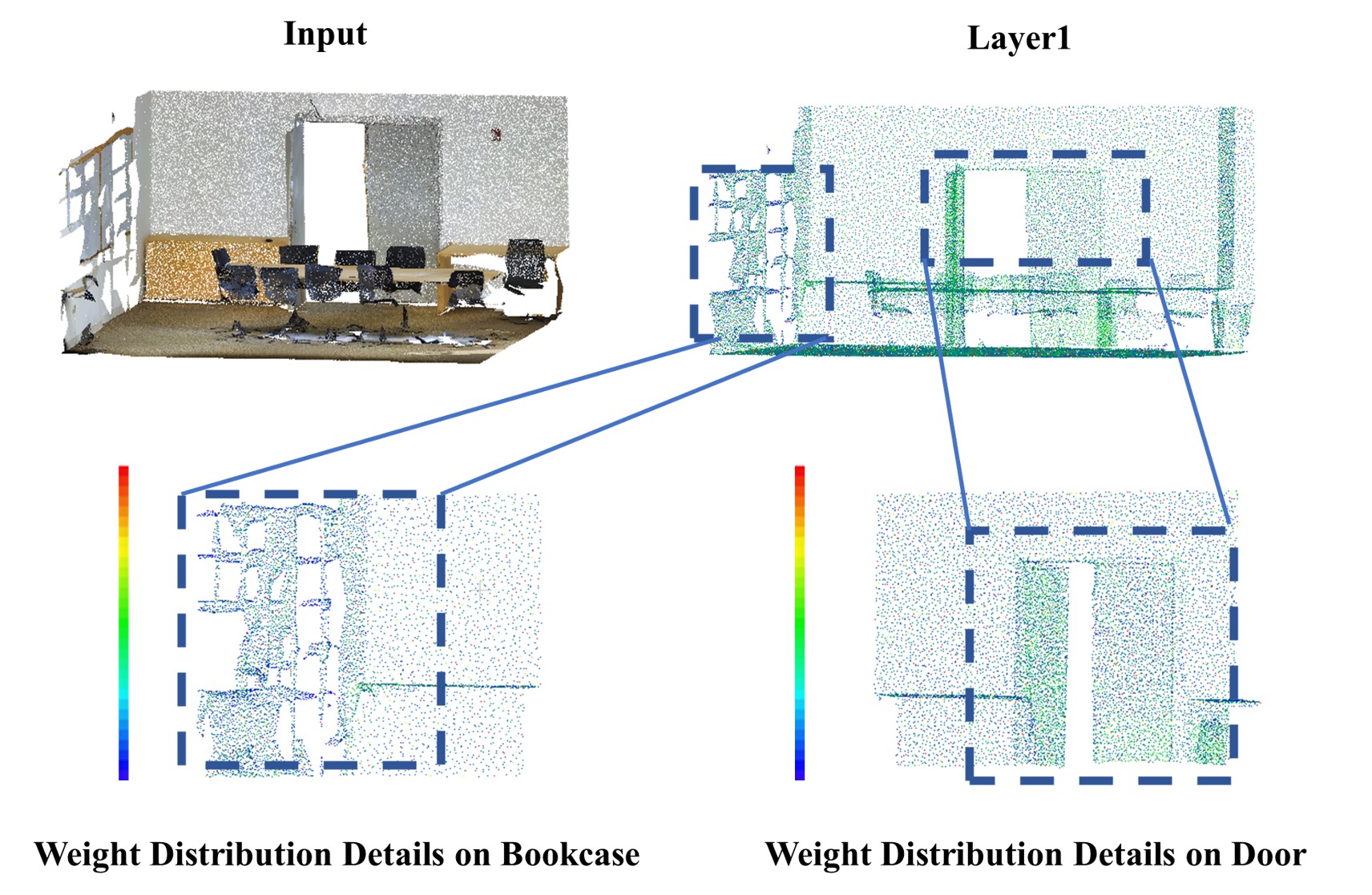}
        \end{minipage}
    }
    \subfigure[]{
        \begin{minipage}[b]{.47\linewidth}
            \centering
            \label{fig:vis sub2}
            \includegraphics[width=1.0\linewidth]{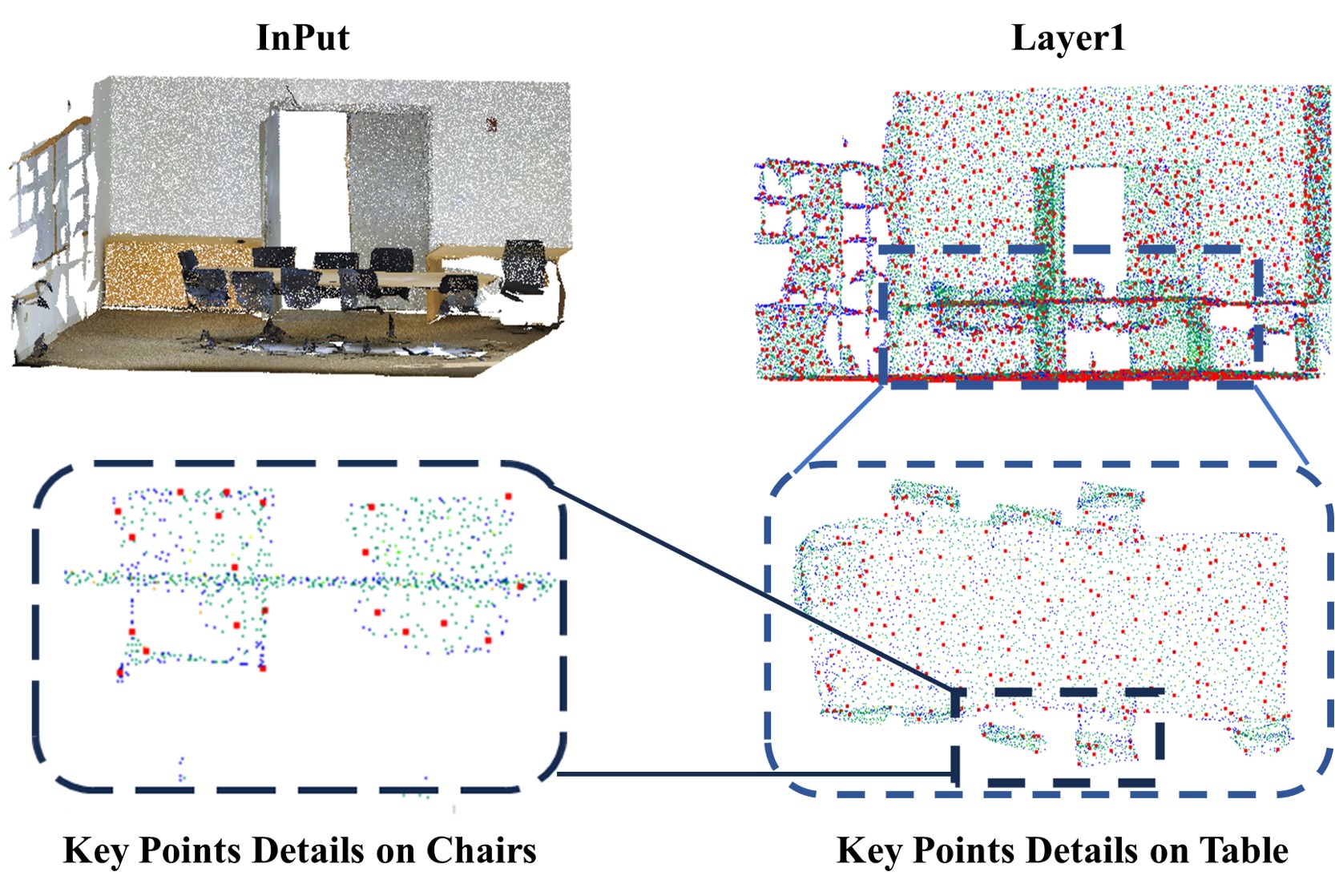}
        \end{minipage}
    }
    \caption{Visualization result of attention distribution. (a) show the visualization details of the corrected attention weight distribution in the first layer; (b) are an enlarged view of some details of the key point allocation in the scene. For the visualization of the key point location, we enlarged the size of the keypoint for easier observation. The color of the points ranges from blue to red, representing the increase in their weights from low to high. The red points are key points.}
    \label{fig:vis result}
    \vspace{-8pt} 
\end{figure*}

The ablation results are summarized in \cref{tab:method of de-nosing}. The result shows that all experiments are higher than baseline model(PointNet++) because of the improvement operations, and $A1$, $A2$ gets lower performance than $A3$ because of the difference between de-nosing method. 

$A1$ uses correlation weights to correct the each line of neighbor feature matrix $f_{N_i}$ which is composed of relative coordinates and distribution features, but as we discussed in \cref{sec:analysis and improvement}, the matrix $f_{N_i}$ should be considered as a whole; directly correcting the structural features of a single point pair will cause the correction features to share the correlation weights with the original features, which reduces the correction effect of $v_{ij}$ on the distribution of features in the high-dimensional space. Therefore, $A1$ gets the lowest result. 

$A2$ avoids this error by directly correcting the point features with correlation weights, but achieves the second best performance due to the lack of noise suppression caused by noise point relative coordinates during neighbor aggregation, which will increase the distribution variance of $f_i$ in the high-dimensional space. These results prove that our de-noising method for irrelevant points is effective.
\begin{table}
\setlength{\tabcolsep}{15pt}
  \centering
  \small
    \caption{Ablation Results of De-nosing method.}
  \begin{tabular}{cccc}
    \toprule
Experiment  & OA & mACC & mIoU \\
    \midrule
A1 &  90.0   &  75.6  & 69.2  \\
A2 &  90.9   &  75.9  & 70.4  \\
\midrule
A3 &  91.0  &  77.4  & 71.2  \\ 
    \bottomrule
  \end{tabular}
  \label{tab:method of de-nosing}
\end{table}
\paragraph{Visual Analysis of Improving Classes Separability}
\label{par:Visual Analysis of Improving Classes Separability}
To further analyze the effect of proposed correction operation CICS, we designed the visualization experiment as shown in Fig.\ref{fig:vis result}. 

We use the correlation weights in PDSA of layer 1 for analysis. We statistically sum the weights of all points in the corresponding neighbors, and show them in the form of heat maps in Fig.\ref{fig:vis result}. Blue to red respectively represent the attention degree of the network, and red points are the key concerns. 

As shown in the Fig.\ref{fig:vis sub1}, regular planes such as walls, door panels, and tabletops are mostly green, which means that the attention of the network is average due to uniform spatial distribution, while edge parts such as bookcase borders are mostly blue that means network does not focus on areas with significant distribution differences. This is consistent with our improvement idea that the weights of PDSA effectively correct the feature distribution variance. 

As shown in the Fig.\ref{fig:vis sub2}, the red points are evenly distributed on the desktop, while on the chair, they are concentrated in the edges and corners, indicating that the network pays more attention to capturing the shape characteristics of the object. key point can be used to represent the neighbor in global attention calculation.

%

\paragraph{Method of Embedding Lightweight Cross-stage Structure Descriptor}
\label{par:Method of Embedding Global Distribution Feature}
We propose LCSD as the supplementary information for refining neighbor structure description; this cross-stage structure descriptor maintains two critical properties: (1) precise encoding of local spatial distributions at each neighbor aggregation, and (2) retains the geometric correlation between points during the cross-stage aggregation process.

Thus, we design the following experiments to compare the performance between different embedding methods of LCSD, and result are shown in \cref{tab:method of embdedding distribution feature}, where $PD$ and $PH$ denote the initial description method which use the octant distribution and  octant centroid respectively, and $D_w$ denotes the using distant relative weight during cross-stage aggregation.
\begin{table}[h]
\setlength{\tabcolsep}{12pt}
  \centering
  \small
    \caption{Ablation Results of LCSD Embedding Method.}
  \begin{tabular}{ccccc}
    \toprule
Experiment  & PD & PH & $D_w$ & mIoU \\
    \midrule
B1 &  \checkmark   &  &  & 70.3  \\
B2 &    & \checkmark &    & 68.5  \\
B3 &    & \checkmark &  \checkmark  & 69.1  \\
\midrule
B4 &  \checkmark  &    & \checkmark & 71.2  \\ 
    \bottomrule
  \end{tabular}
  \label{tab:method of embdedding distribution feature}
\end{table}

In \cref{sec:analysis and improvement}, we calculate the correlation between neighboring point and neighbor to correct the ${f}_{{N}_{i}}$ matrix, which needs to describe the aggregated structure. The results in \cref{tab:method of embdedding distribution feature} intuitively shows that using octant distance designed by us as spatial feature can achieve a better performance of more than $2\%$ mIou improvement regardless of whether distant weight is adopted. This means the encoding method using distant weight can better retain relevant information after dimensional compression. 

Moreover, all experiments can improve performance though distant relative weight $D_w$, and $B3$, $B4$ improve the mIou by $0.6\%$, $0.9\%$ respectively; this result demonstrates that $D_w$ not only enhances feature correlation preservation during cross-stage aggregation but also validates our analysis: leveraging global point correlations to optimize high-dimensional feature distributions in neighbor aggregation significantly improves the network’s representational capacity.

\paragraph{Dimension of Additional Cross-stage Spatial Descriptor}
\label{par:Dimension of Global Distribution Feature}
The PDSA we proposed corrects the high-dimensional spatial distribution of features in the point cloud neighbor aggregation process based on the correlation of global distribution. The basis of this process is the additional cross-stage spatial descriptor $d_i$. We have already discussed the encoding method of this feature. 

However, since this feature requires information compression when aggregated across stages to avoid the problem of dimension explosion, we have designed the following experiments to explore the optimal compression dimension. The experimental results are shown in Tab.\ref{tab:ablation dimension of global distribution feature}.

\begin{table}
\setlength{\tabcolsep}{6pt}
  \centering
  \small
    \caption{Ablation Results of Additional Cross-stage Descriptor Dimension.}
  \begin{tabular}{ccccc}
    \toprule
Dimension  & OA & mACC & mIoU & param(M)\\
    \midrule
1 &   89.5  &   76.1 & 69.2 & 8.5 \\
2 &  90.72   &  75.67  & 69.43 & 11.2 \\
    \midrule
3 &  91.0   &  77.4  &  71.2 & 12.5 \\
    \midrule
4 &  91.2  & 77.4   &71.3  &21.4 \\
    \bottomrule
  \end{tabular}
  \label{tab:ablation dimension of global distribution feature}
\end{table}

Through the experimental results in Tab.\ref{tab:ablation dimension of global distribution feature}, we can intuitively see that as the compression dimension increases, the performance of the network is also improving. 

However, we can also observe that due to the  computational complexity of attention mechanism is in the form of $O(N^2)$, the parameters and computational cost increase exponentially with the increase of dimensions, and the resulting performance improvement is not significant. Through experimental comparison, we chose 3 as the best compression dimension, which is also consistent with the setting in PointHop \cite{pointhop}.

\section{Conclusion}
In this paper, we analyze the noise problems existing in the point cloud neighbor aggregation module represented by SA module: (1) noise interference of irrelevant points during the neighbor aggregation process and (2) confusion of classes distribution caused by hierarchical gap. To solve above problems, we propose corresponding solution example: PDSA module, which corrects the high-dimensional spatial distribution of features based on the correlation of lightweight cross-stage structures, avoiding the huge computational overhead and noise sensitivity caused by directly using low-dimensional geometric structures for improving neighbor description.

We conduct experiments on semantic segmentation and classification tasks based on different network baselines on three well-known datasets; the results achieve performance and parameter efficiency. PDSA is the initial attempt of the problems we have analyzed. In the future, we will further explore the best solution to this problem, especially in further improving the local detail processing optimization of the large-scene point cloud.

\section*{Limitations}

A key limitation of ASCD is its incompatibility with FlashAttention. Because our method must dynamically modify the attention matrix at inference time, it cannot make use of the fused kernels, leading to higher memory consumption and slower decoding.
A promising workaround is to distill the steering signal into the model during training: we can add an auxiliary loss -- e.g., a KL-divergence term -- that drives the native attention distribution to approximate the ASCD target distribution. If successful, the model would internalise the hallucination-mitigation behaviour, removing the need for on-the-fly edits and restoring FlashAttention speed-ups. We regard training-time attention regularization as a promising direction: it could distill the hallucination-robust behaviour discovered by training-free, attention-modified methods into the model itself, so that at inference the model retains this robustness while fully benefiting from FlashAttention’s speed and memory efficiency.



\bibliography{main}

\end{document}